\documentclass[pdflatex,sn-nature]{sn-jnl}

\usepackage{graphicx, stfloats, float, xr-hyper}
\usepackage{caption}
\usepackage{tabularx}
\usepackage{amssymb, amsmath, standalone, comment}
\usepackage{subfiles}
\usepackage{multirow}
\usepackage{lscape}
\usepackage{booktabs, tabulary, array, colortbl}
\usepackage{xcolor}
\usepackage{rotating}
\usepackage{comment}
\usepackage{pifont}
\usepackage{pdflscape}
\usepackage{afterpage}

\usepackage{amsfonts}%
\usepackage{amsthm}%
\usepackage{mathrsfs}%
\usepackage[title]{appendix}%
\usepackage{textcomp}%
\usepackage{manyfoot}%
\usepackage{booktabs}%
\usepackage{algorithm}%
\usepackage{algorithmicx}%
\usepackage{algpseudocode}%
\usepackage{listings}%
\usepackage{rotating}
\usepackage{longtable}

\usepackage[switch]{lineno}

\usepackage[nolist,nohyperlinks]{acronym}
\usepackage{orcidlink}

\usepackage{hyperref}
\hypersetup{
    colorlinks=true,
    linkcolor=blue,
    citecolor=teal,
    filecolor=magenta,      
    urlcolor=cyan,
    pdftitle={main},
    pdfpagemode=FullScreen,
    }
\definecolor{TC}{gray}{0.95}

\begin{acronym}
    \acro{CNN}{Convolutional Neural Network}
    \acro{SVM}{Support Vector Machine}
    \acro{MLP}{Multi Layer Perception}
    \acro{FBG}{Fiber Bragg Gratings}
    \acro{BRISK}{Binary Robust Invariant Scalable Keypoints}
    \acro{CoM}{Center of Mass}
    \acro{MI}{Minimally Invasive}
    \acro{IMU}{Inertial Measurement Unit}
    \acro{LSTM}{Long Short-Term Memory}
    \acro{ANN}{Artificial Neural Network}
    \acro{DNN}{Deep Neural Network}
    \acro{kNN}{k-Nearest Neighbours}
    \acro{CPER}{Contact Prioritized Experience Replay}
    \acro{RL}{Reinforcement Learning}
    \acro{RNN}{Recurrent Neural Network}
    \acro{PCA}{Principal Component Analysis}
    \acro{RMSE}{Root Mean Square Error}
    \acro{RF}{Random Forest}
    \acro{STAG}{Scalable Tactile Glove}
    \acro{GRU-AE}{Gated Recurrent Unit-base AutoEncoder}
    \acro{LDS}{Linear Dynamical System}
    \acro{VLAD}{Vector of Locally Aggregated Descriptors}
    \acro{SA}{Slowly Adapting}
    \acro{MPC}{Model Predictive Control}
    \acro{HMM}{Hidden Markov Model}
    \acro{LfD}{Learning from Demonstration}
    \acro{POMDP}{Partial Observable Markov Decision Process}
    \acro{DoF}{Degree of Freedom}
    \acro{PDMS}{PolyDiMethylSiloxane}
    \acro{PVDF}{PolyVinyliDene Fluoride}
    \acro{DWT}{Discrete Wavelet Transform}
    \acro{FIFO}{Finger-skin Inspired Flexible Optical}
    \acro{EGaIn}{Eutectic Gallium-Indium}
    \acro{IC}{Integrated Circuit}
    \acro{CNT}{Carbon NanoTube}
    \acro{PAN}{Poly-AcryloNitrile}
    \acro{EIT}{Electrical Impedance Tomography}
    \acro{TDNN}{Time-delay Neural Network}
    \acro{MSE}{Mean Square Error}
    \acro{RL}{Reinforcement Learning}
    \acro{BoW}{Bag of Words}
    \acro{DMP}{Dynamic Movement Primitive}
\end{acronym}

\begin{document}
    \title[Sensorimotor Control Strategies for Tactile Robotics]{Sensorimotor Control Strategies for Tactile Robotics}
    
    \author*[1,2]{\fnm{Enrico} \sur{Donato}}\email{enrico.donato@santannapisa.it}
    \author[3]{\fnm{Matteo} \sur{Lo Preti}}
    \author[4]{\fnm{Lucia} \sur{Beccai}}
    \author*[1,2]{\fnm{Egidio} \sur{Falotico}}\email{egidio.falotico@santannapisa.it}
    
    \affil*[1]{\orgdiv{The BioRobotics Institute}, \orgname{Sant'Anna School of Advanced Studies}, \orgaddress{\city{Pontedera}, \postcode{56025}, \country{Italy}}}
    \affil*[2]{\orgdiv{Department of Excellence in Robotics and AI}, \orgname{Sant'Anna School of Advanced Studies}, \orgaddress{\city{Pisa}, \postcode{56125}, \country{Italy}}}    
    \affil[3]{\orgdiv{Advanced Robotics Centre}, \orgname{National University of Singapore}, \orgaddress{\city{Singapore}, \postcode{117575}, \country{Singapore}}}
    \affil[4]{\orgname{Italian Institute of Technology}, \orgaddress{\city{Genova}, \postcode{16163}, \country{Italy}}}
    
    \abstract{
        How are robots becoming smarter at interacting with their surroundings? Recent advances have reshaped how robots use tactile sensing to perceive and engage with the world. Tactile sensing is a game-changer, allowing robots to embed sensorimotor control strategies to interact with complex environments and skillfully handle heterogeneous objects. Such control frameworks plan contact-driven motions while staying responsive to sudden changes. We review the latest methods for building perception and control systems in tactile robotics while offering practical guidelines for their design and implementation. We also address key challenges to shape the future of intelligent robots.
    }
    
    
    \maketitle

    \section{Introduction}
        Physical contacts are at the base of each embodied interaction. As for living beings, also robots continuously establish diverse contacts to fulfill their tasks. 
        
        Over the last decades, one of the bold goals of robotics research has been to provide artificial agents with dexterity and adaptability - typical of biological systems - while interacting with their surroundings. Despite the huge work and the excellent outputs in this field, such capabilities still require hard refinements and studies to be fully delivered on our robots. 
        
        The scientific contribution to this objective builds upon three pillars: the design of an appropriate \textit{embodiment} - concerning its morphology, actuation strategy, and sensing technology; feature extraction algorithms from tactile signals to build a \textit{perception} model of the experience; closed-loop robot \textit{control} strategies that drive robot decisions according to either raw tactile feedback or perceptual representations. The contribution of our paper builds upon the last two pillars, leveraging the exhaustive discussion of embodiments of further sources \cite{dahiya2013,xi2024,park2024}, alongside considerations in the following sections due to the interleaved nature of the challenge. We specifically provide an extensive review of methodologies that allow for feature extraction from tactile sensory feedback, and discuss how they could be employed within ad-hoc controllers to provide general design guidelines.
        
        The challenge of getting a fully flourished \textit{tactile robotics} can be explained by looking at the difference between touch and other well-defined senses. Biological touch lacks of a localized sensory organ, forcing researchers to select some of the pertinent features and modalities to be artificially represented \cite{dargahi2004}. For example, the human hand senses outer stimuli that cause mechanical transients in its inner tissues (e.g., indentation, vibration) by mechanoreceptors \cite{vallbo1976,johansson1983}, and the adaptation rate of its skin underscores the need for tactile sensors that generate high temporal-varying contact forces. Furthermore, although spatial distribution is widely recognized as a crucial aspect of tactile perception, existing literature has predominantly focused on technologies and applications that rely on localized sensing \cite{chi2018}. Therefore, tactile sensors have been typically developed as discrete sensing units, to produce robust and localized information. Anyway, the most ambitious objective lies in the development of fully distributed tactile skins \cite{shih2020}. Consequently, robot perception and control must account for temporal-varying, spatially distributed signals over different features and modalities. In addition, the information resulting from variable environmental conditions presents a significant challenge \cite{wan2016}, yet contributes to striking a balance between advanced sensory systems and resilient controllers.

        \begin{figure*}[t]
          \includegraphics[width=\textwidth]{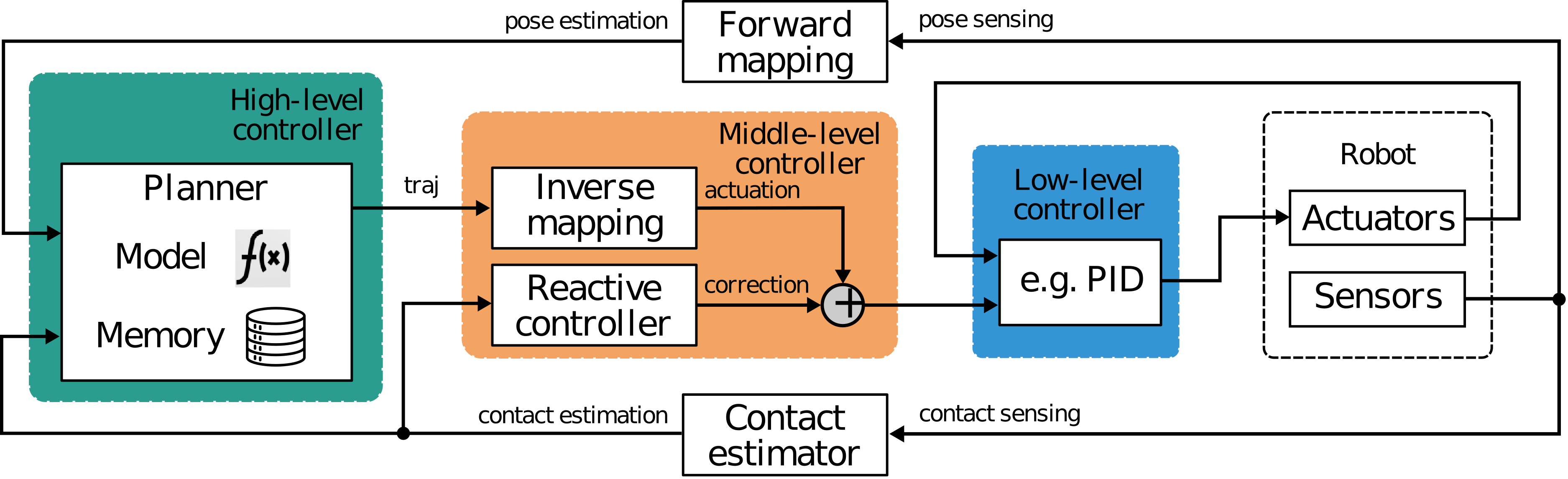}
          \caption{Sensors measure the robot's state and contacts with the environment, through a forward mapping and a contact estimator respectively. Such information is primarily sent to a reactive controller, which can generate a motion correction command for the robot to quickly respond to control stability or robustness losses when contact is achieved. The command is sent to the robot's actuation system through low-level controllers. Similarly, post-processed sensing information is sent to a high-level controller, which allows to plan the manipulation task, making also use of robot models and a memory of past system states. The trajectory is then sent to an actuation command generator specific to the robotic system under examination, which is located within a mid-level block and is added to the previously generated correction.}
          \label{fig:control_architecture}
        \end{figure*}
        
        
        This evidence pushes towards the choice of \textit{sensorimotor control architectures} to integrate sensory information with motor control - generically represented in Figure \ref{fig:control_architecture}. The implementation of a reactive controller (orange) and a planner (green) represents respectively the use of either raw sensory feedback or perception models. The former operates on sets of heuristic rules to generate predefined routines and the latter builds a model of the sensory experience - possibly recurring to a memory storage - and provides a representation that best summarizes the sensory observation. Planning is thus crucial for implementing interaction strategies, allowing for informed decisions and generating action sequences using models \cite{vanhoof2015, palleschi2022manipulationplanning}. Predictive control algorithms, for instance, utilize such models to predict the system behavior and optimize control actions over time \cite{tian2019}. Despite these advantages, working only over perception models might induce temporal delays intrinsic to the computational burden they introduce. This might result in control failure of the system, like the full slippage of an object which happens at a very high time rate. Thus, reactive control comes as a solution to adapt to unforeseen situations in a short time and maintain safe and reliable interaction with the environment \cite{bohg2017interactiveperception}, like when a robot encounters an obstruction while manipulating an object and triggers an immediate response to either adjust the contact force or change the motion trajectory \cite{alshuka2018impedance}. These two fundamental components - reactive controllers and planners - will be further detailed, and their roles will be color-coded in proposed control architectures.

\begin{figure*}[t]
          \includegraphics[width=\textwidth]{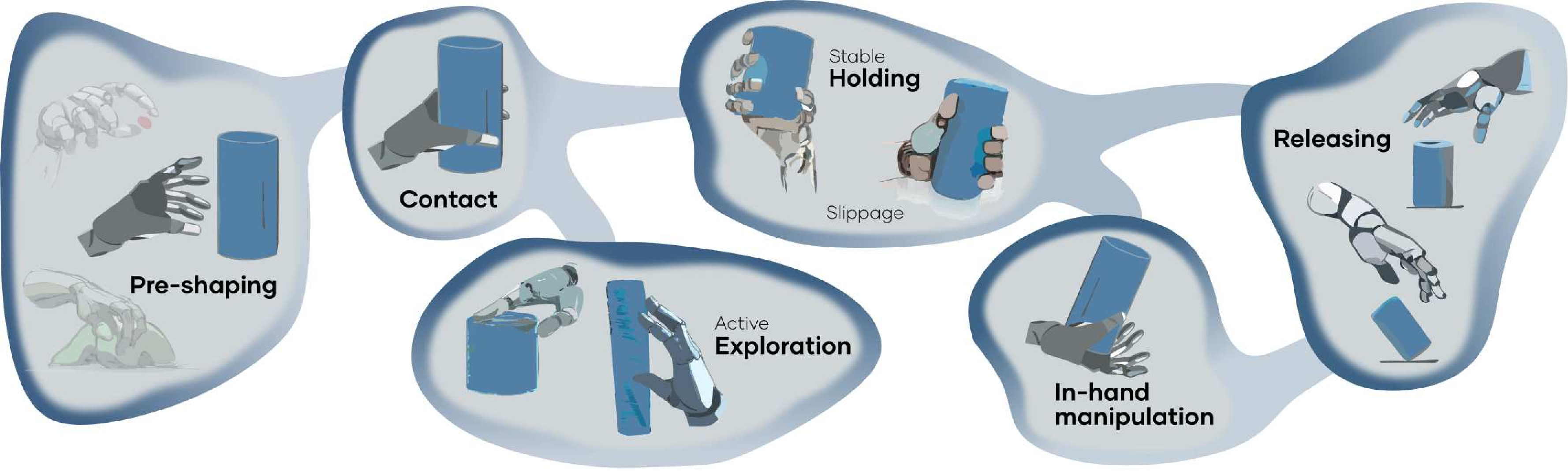}
          \caption{The manipulation process involves several distinct stages. Before initiating the grasp, the gripper adjusts its fingers to facilitate future contact sensing and explores the objects in its surroundings. Once a secure grasp is achieved, the gripper firmly holds the object to prevent slippage and enables in-hand manipulation. Finally, the object is placed onto a designated target configuration.}
          \label{fig:grasping_stages}
        \end{figure*}

        The gold standard of sensorimotor control is the manipulation capability of the human hand. Therefore, we usually refer to human hand strategies to classify interactions and devise variations in sensorimotor control implementations. As evident from Figure \ref{fig:grasping_stages}, the manipulation process can be divided into different stages - even though not all of them might occur - and coordinated by contact events derived from tactile sensing \cite{yamaguchi2019}. Localization, positioning, and object pickup occur during the initial grasp. As the task progresses, tactile sensors - alongside the visual system - become essential to measure contacts with objects \cite{marwan2021}. Such tactile perception becomes particularly important in unstructured environments, multi-object grasping, and planning for in-hand manipulation. The final stage involves releasing the object at the intended location \cite{wang2020}. Stages are represented as distinct and sequential for ease of explanation, but interleaved execution can arise in real-world applications.

        This paper aims to review sensorimotor robot control architectures to handle different manipulation stages - \textit{active exploration}, \textit{grasping}, and \textit{dexterous manipulation} \cite{dahiya2010}. We first detail the nature of tactile sensory feedback and propose some desiderata for designing future tactile sensors and body sensorization to extract more informative outputs for control architectures. Later, we will explore single-stage control architectures to highlight current drawbacks and provide indications for future implementations.
        

    \section{Tactile Sensing: Feedback for Interaction} \label{section:sensing}
        Contacts are at the heart of any robot interaction, which makes their perception crucial to implement closed-loop control strategies. Control requirements can be set as desiderata to guide tactile sensor design and body sensorization, and to drive feature extraction to enhance robot perception.

        \subsection{Tactile sensors: multi-sensing or multi-modality?}
        

        Tactile sensorization implies a perceptual decoupling between internal and external stimuli to stabilize the interaction locally and have interpretable tactile information to control both the contact and object states \cite{hogan2020,zhang2020a}. Modalities arising from tactile signals comprise contact, pressure distribution, slip, vibration, and temperature \cite{yamaguchi2019}, usually led by a task-driven choice.

        Even though a modality-complete tactile sensor does not exist, alternatives are either multi-modal discrete sensors or multiple sensors each detecting a different stimulus. For instance, a multi-label detection model that takes tactile sequences as input and simultaneously predicts the object stiffness, thermal conductivity, roughness, and texture has been proposed \cite{han2018}, as well as a multilayer pressure and temperature sensor \cite{fastierwooller2021}. Similarly, an ultrasensitive self-powered multimodal sensor was made by exploiting triboelectric and pyroelectric effects to measure pressure and temperature in real-time \cite{shin2022}.
        
        Multiple sensing elements are usually combined into arrays to sensorize large areas, such as a hand palm or a torso. A single large-area sensor is still a low explored solution since any trivial correspondence between the sensing element and the sensing point is lost. Obtaining a distributed sensing map hence implies solving a non-linear reconstruction problem \cite{chen2022large}. For example, a pressure map reconstruction is obtained by Lo Preti et al. \cite{lopreti2022} on a large-area optical sensor, in which the polymeric substrate of the sensor acts as a continuum waveguide, while photo emitters and receivers are located at the boundaries, making the sensing area free from electronic components. Alternatively, Hardman et al. \cite{hardman2023eit} propose the combination of functional stretchable materials with Electrical Impedance Tomography as a promising method for developing continuum sensorized robotic skins with high resolutions. All these alternatives are corroborated by advances in processing techniques, used to retrieve tactile data from complex configurations. Indeed, embedded local and distributed tactile data processing is centric in large-area sensors \cite{soni2020}.

        \subsection{Information uncertainty and real-time feature extraction}    
        Sensor selection is driven according to yet-known target features, as control strategies designed for objects with known features cannot be applied directly under feature uncertainty. For example, a gripper needs to localize a target and evaluate the affordance before any holding; later, the object can be grasped and manipulated to accomplish the desired task. Tactile information could be exploited for this case and to provide a better estimate of the physical characteristics of grasped objects \cite{jiao2020}, object feature identification with finite grasp sets \cite{steffen2007,takamuku2008,bykerk2019}, from noisy signals \cite{ratnasingam2011} or with exotic sensors \cite{zhang2018}. 
        
        Targets can be identified as \textit{geometric-} or \textit{physical-uncertain} if there is limited knowledge of either the target's geometry and global pose, or its inertial properties. If both are uncertain, objects are considered \textit{unknown}. We propose a gross indication for the respective tactile sensorization at each manipulation phase in Table \ref{Table:Sensing}. During the reaching phase, a proximity sensor modulates the acceleration until contact is made. After contact, sensors estimate properties like pose - by probing multiple points - and texture \cite{guo2024vibro}. The object pose helps select the best gripper position for an effective grasp, while texture and stiffness adjust the initial force based on the estimated friction. While grasping, tactile information becomes more important than vision. The applied force is constantly monitored to avoid damage, detect sudden changes, and use shear force monitoring to secure the grasp. For releasing, a proximity sensor guides the gripper to the target area, and interaction with the surface creates vibrations detected by a slip sensor, and a contact sensor confirms the release.


        \begin{table}[b]
            \centering
            \begin{tabular*}{.9\textwidth}{@{\extracolsep\fill}lcccc}
                \toprule
                & \multicolumn{1}{c}{Geometry-uncertain} & \multicolumn{1}{c}{Physical-uncertain} & \multicolumn{1}{c}{Unknown object} \\
                \midrule
                Reach & \multicolumn{3}{c}{Proximity, contact} \\
                \midrule
                Contact & Force only & \multicolumn{2}{c}{Texture, stiffness, force} \\
                \midrule
                Hold & \multicolumn{3}{c}{\cellcolor[HTML]{e2e2e2}Force, slip} \\
                \midrule
                & \multicolumn{3}{c}{\cellcolor[HTML]{e2e2e2}Contact} \\
                \multirow{-2}{*}{Release} & \multicolumn{3}{c}{Proximity, slip} \\
                \bottomrule
            \end{tabular*}
            \caption{End-effector sensorization to accomplish a prehensile task. The four stages of the task are: reach, contact, hold, and release. Sensors configuration changes if geometry is known, and vision (grey cells) can contribute.}
            \label{Table:Sensing}
        \end{table}
        
        Eventually, tactile data should inform \cite{tegin2005} on:
        \begin{enumerate}
        	\item the presence of contacts with the target;
        	\item which are global target features - surface, edges, pose;
        	\item which are local mechanical features - texture, compliance, friction, pressure, contact forces, torques, slip, vibrations, temperature, moisture.
        \end{enumerate}
        In addition, properties like the center of mass or the weight can be estimated by actively manipulating the target. 
        
        
        For geometry-uncertain objects, position and shape must be estimated. The most common solution is to use vision, while tactile perception is a different way to address this problem: a blind search with contact sensors is used to determine the position of the target, and the shape can be inferred by maximizing local information or minimizing exploration costs.
        
        In the case of physically uncertain objects, inertial properties - mass and stiffness - are considered. Mass is usually detected with force-torque sensors \cite{kubus2008}, or calculated from the 3D force vector of tactile sensors at fingertips \cite{silva2019}. Such estimation is driven by active perception, thus applying a variable set of contact forces to inform the estimation and refining the search space according to the recorded evidence, paying attention not to apply forces that result in unsuccessful or destructive grasps. Another key state is played by the object's internal state, which could be inferred from high-frequency tactile information, for example to detect liquids in containers \cite{chitta2011}.
        

        Stiffness is instead commonly formulated as a classification problem: despite an approach that is coupling motion estimation and optic flow \cite{zang2010}, it usually relies on tactile sensors \cite{drimus2014}. Manipulating objects of unknown stiffness is challenging due to unpredictable deformations \cite{frank2014}. This is addressed by categorizing objects based on deformation characteristics \cite{luo2017}, and employing optical sensors to detect elastic object deformation during manipulation \cite{mankowski2020}. Materials can be classified by analyzing force oscillations or gradual increases through their "deformability degree" \cite{delgado2017b}. We can estimate the stiffness, by analyzing force curves from tactile arrays with sensorized gripper-object interaction, thus facilitating material discrimination \cite{ji2019}.
        For deformable object manipulation, piezoelectric sensors prove effective, and capable of extracting wide-range stiffness and damping coefficients, particularly useful for Minimal Invasive Surgery \cite{esmaeel2020}.         
        
        Feature extraction may limit real-time use as feedback for sensorimotor controllers. Processing high-dimensional information can be computationally demanding, though redundant, body-distributed data can enhance grasping strategies \cite{ascari2009}. Addressing this issue involves both hardware and software improvements. Advances in tactile sensor fabrication, such as \ac{STAG} with numerous sensitivity points, support big-data tactile exploration \cite{sundaram2019}. Real-time object classification \cite{li2020a} illustrates how tactile information from multiple sources can identify material, shape, and size with low latency by integrating spatial and temporal data.

        \subsection{Body sensorization for interaction}
        Along with variability in sensing modalities and their temporal dynamics, embodied sensor distribution spatially shapes the information feedback and might alter its transient.
        
\begin{figure}[b]
            \centering
            \includegraphics[width=.7\textwidth]{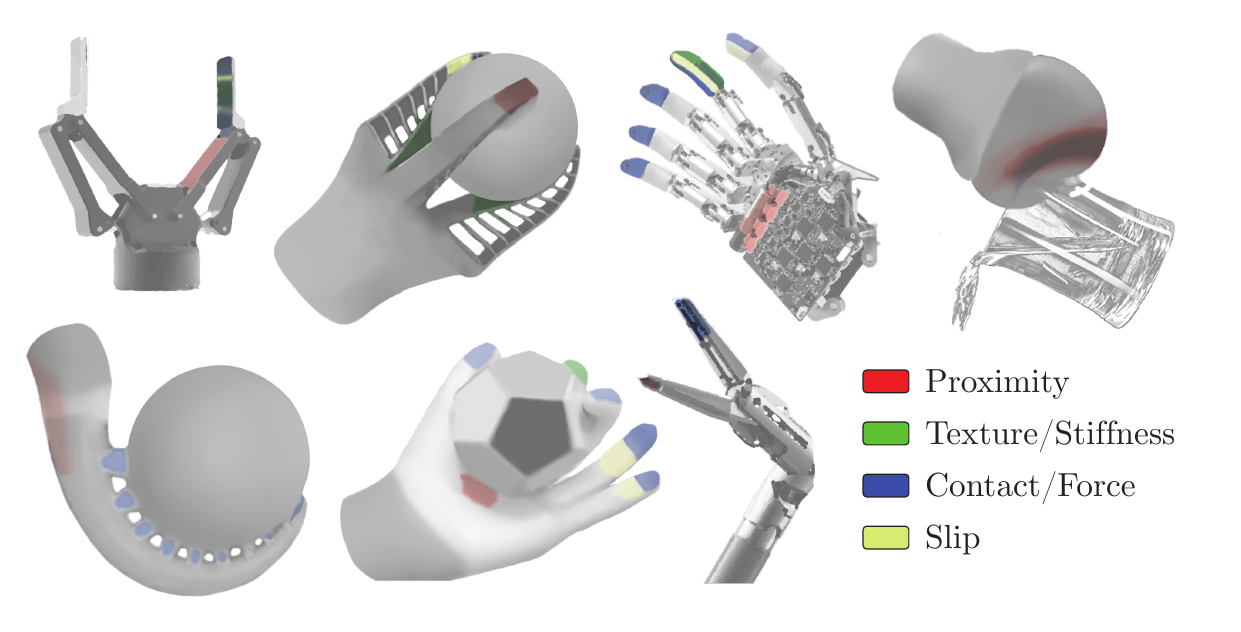}
            \caption{Tactile sensorization of robotic end-effectors: two-fingered, fin-ray, anthropomorphic, tentacle, and universal. Areas devoted to different sensing have been highlighted with different colors: proximity (red), target properties - texture, stiffness, etc. - (green), pressure and force (blue), and slip (yellow).}
            \label{fig:GripperSensorization}
        \end{figure}

        Our discussion mostly focuses on interactions by end-effectors, such as artificial hands and grippers. Figure \ref{fig:GripperSensorization} showcases how different modalities should be spatially perceived for each embodiment to carry on manipulation tasks. Concerning anthropomorphic hands \cite{mendez2021current}, proximity sensors can be integrated into the palm while the forefinger is predominantly used for surface exploration and should contain texture and stiffness sensors as well as force/contact and slip sensors \cite{lederman1988physiology}. Force/contact and slip sensors are also helpful in the thumb during pinching \cite{controzzi2018progress, romeo2020methods}. The role of the remaining fingers during prehension tasks is of secondary importance, while tactile information is still useful during dexterous manipulation tasks. If two- and three-finger grippers are selected, palpation and grasping are done with the internal part of the fingers. Hence, texture/stiffness and force sensors should be placed along fingers since these deformable parts conform to the object \cite{li2023design}, whereas slip is measured at fingertips \cite{liu2022gelsight}. Proximity sensors are placed on the dorsal side for exploration. 

        Deformable embodiments undergo distributed deformations during interactions, allowing for improved versatility and adaptability. Yet, deformable body sensorization turns into a more challenging task than traditional, even if their level of dexterity does not require a finer feature recognition. Examples of deformable embodiments are tentacles and universal grippers: in the former case, proximity can be detected at the outer part of the, while force can be measured at the tip to indicate the adherence of suckers \cite{zhuang2018fbg, frey2022octopus}; for the latter, proximity sensors can be placed closer to the edge, while forces should be measured in the middle \cite{hughes2018tactile,loh20213d}. 

        
        Hence, the process of sensorization determines the purpose of tactile sensors and their body placement for specific manipulation tasks. Thus, it empowers: 
        \begin{itemize}
        	\item adaptation to the environment;
        	\item greater ability to pick up objects \cite{homberg2019} with different shape, texture, and stiffness \cite{tawk2019}; 
        	\item simpler grasp planning with finer motions \cite{bogue2017};
        	\item exertion of small gripping force for object manipulation;
            \item counterbalance of uncertainties inherent in deformable bodies, like their compliance and friction.
        \end{itemize}
        
        
        
        \subsection{Limitations and current bottlenecks}
        The variety of tactile sensors in the literature is exponentially growing, and so is the number of sensorized grippers \cite{shintake2018, park2020, goh20223d, zhang2022nondestructive, roels2022self}. Multiple patches applied on rigid systems are a reliable solution, and wirings can be routed with minor issues \cite{schmitz2011, mittendorfer2011}. 
        
        The main drawback of rigid embodiments is their lack of gentle and self-interaction with an unknown environment. Soft embodiments come as a solution, but their morphology poses challenges since rigid sensors cannot conform and adapt to complex shapes. The advent of hybrid and soft grippers raised new sensorization challenges \cite{wang2018}. Soft tactile sensors in a compliant architecture will lack anchor and reference points, making it difficult to decouple between self- and external-induced deformations. Similarly, material degradation inherently leads to non-stationary behaviors, like baseline drifts. These effects are undesirable in sensors, however they are often ignored or treated as unmodeled noise \cite{rolf2015}. Alternatively, data-driven algorithms use signal history to encode time dependency for dynamic signals. This process can be explicit when using time-dependent machine learning algorithms, or implicit if multiple delayed inputs are concatenated as input of the model \cite{chin2020}. Additionally, time adaptive models or periodical re-calibration \cite{rolf2015} can alleviate the problem. 
        
        When multiple soft tactile units are combined, the high number of connections may introduce undesired mechanical constraints or an uneven stiffness distribution in compliant systems \cite{lopreti2020}. Moreover, the computational workload rapidly grows with the number of sensors. Due to sparse activations, this problem may be faced by creating a hierarchical processing unit to compress data or relying on event-based preprocessing \cite{afshar2020}.
         
        Thus, the most common bottlenecks in sensorized grippers are conformability to the gripper shape \cite{hughes2020}, calibration and decoupling \cite{park2020}, signal hysteresis and non-stationarity \cite{duran2012, han2020}, wirings, and high computational burden \cite{alagi2020}.

    \section{Tactile Exploration}
        \label{section:exploration}
        Action planning exploits the robot workspace, which may be unknown or subject to continuous modifications due to robot or environmental changes. Thus, robots must actively gather perception from their surroundings to build a virtual model. Control algorithms that incorporate tactile feedback are crucial for effective workspace exploration.
        
        
        If combined with contact estimation, the exploration strategy may end up in tactile servoing. It ensures that the robot maintains contacts while smoothly sliding along surfaces. Sensor models bridge the gap between sensory information and contact state estimation, despite the quality strictly relies on tactile sensors performance.

        \subsection{Active exploration}
        A trivial workspace exploration is performed by randomly moving the end effector to obtain the target characteristics without any control feedback \cite{prescott2011, yi2016}. This method is less effective and more time-consuming than its active equivalent, which drives action based on sampled data. 

        \begin{figure}[b]
            \centering
            \includegraphics[width=.7\linewidth]{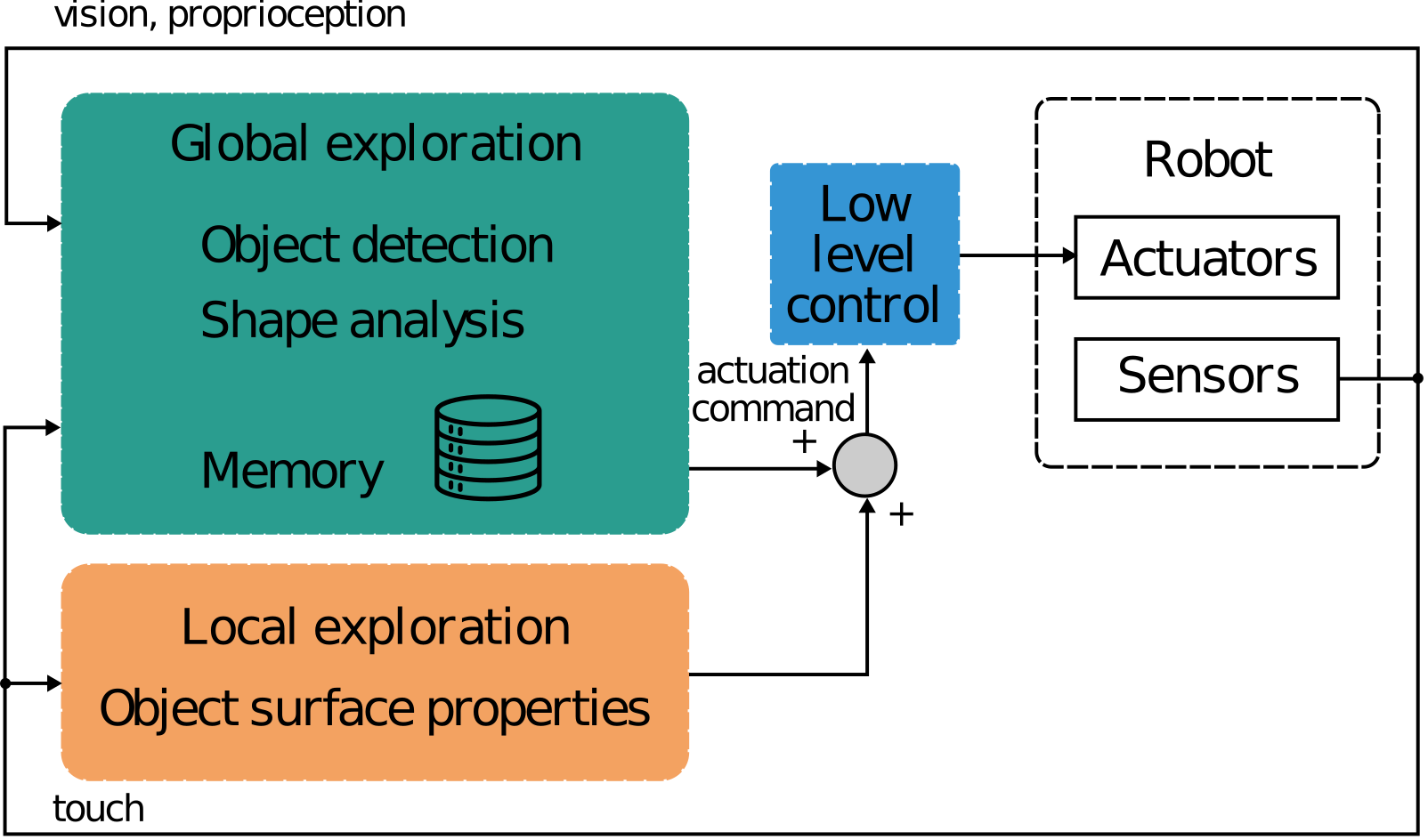}
            \caption{Local and global workspace exploration. Local exploration relies on tactile sensing to gather object surface properties - like its curvature, texture, and roughness. Combined with vision and proprioception, it facilitates higher-level perception like object detection and shape analysis. Global exploration utilizes previously acquired information to make perception-informed decisions, such as object recognition and spatial reasoning.}
            \label{fig:active_exploration}
        \end{figure}

        As presented in Figure \ref{fig:active_exploration}, tactile exploration can be performed \textit{locally} and \textit{globally}. Local exploration is enabled by extrinsic tactile sensors mounted at or near the contact interface. They generate tactile maps that could be used to derive valuable manipulation parameters \cite{li2020b} - like image orientation, edge detection, contact location, grip stability. Alternatively, surface reconstruction has been proposed as a multi-modal sensory fusion problem, where tactile sensing is used to complement RGB-D information in case of occlusions \cite{saund2021,suresh2022,rustler2022}. Lastly, Li et al. \cite{li2014a} proposed camera-based tactile sensors to generate high-resolution tactile images, and their comparison through keypoints localization. Tactile images can track spatio-temporal evolution, allowing for feature extraction for learning-based methodologies \cite{fang2022}. Moreover, tactile images inform human grasping-intention detection while interacting with the robot \cite{li2021}, later translated to robot-object interaction to optimize the reaching-for-grasping phase. Anyway, all these applications are limited to the observation of small contact areas, requiring the acquisition of multiple tactile images that are fused to generate wider observation \cite{heidemann2004,chen2023sliding}.  
        
        Despite their advantages, extrinsic sensors are prone to cross-talk phenomena that might affect tactile image generation and limit the extraction of object features. An adaptive thresholder permits to overcome this effect \cite{berger1991}. A side effect is the removal of valid tactels - as per high force gradient, and low force values. Despite this, it is suggested to maintain a midrange of forces on the sensor during dynamic edge extraction, less sensitive to cross-talk. 
        
        Contrary to local exploration, global exploration cannot solely rely on tactile sensing, as it cannot extrapolate from local data. Consequently, it is fused with either proprioception or vision \cite{salazar2022contour}. iCLAP \cite{luo2019} works as an iterative method to link the kinaesthetic cues and tactile patterns, to recognize object shapes. Similarly, object shapes are reconstructed by combining contact geometry estimation and sliding shape exploration \cite{gupta2022}, even under object motion \cite{aquilina2024dynamicfollowing}. On the same line, online tactile surface exploration of unknown objects with a robotic hand is generated by actively maximizing the information entropy while dynamically balancing the global knowledge and local complexity \cite{khadivar2023}.

        By exploiting both local and global exploration, tactile exploration can be implemented as a behaviour-based task through curiosity-driven approach, to get a set of tactile exploratory skills in the absence of an explicit teacher signal \cite{pape2012}. The acquisition of tactile-driven skills aims to represent gathered data using fewer computational resources and leads to the observation of areas less explored. It consider unknown or unexplored data, and the enhancement of learning capabilities by exploiting new patterns.
        
        Active tactile exploration and learning strategies based on intrinsic motivation are combined in \cite{vulin2021}. The \ac{RL} algorithm works on a reward function that considers the task objective - pick and place, sliding and pushing - and the contact with the target. This method is prone to be generalized to multi-modal sensory systems.

        \subsection{Tactile servoing}
        Tactile servoing - or, sliding over unknown surfaces - is essential for workspace exploration. It requires a model to link sensing and contact states to close the loop and drive future actions. For both its nature and high-frequency demand, it must be conveyed by touch and not vision.

        Tactile and visual servoing differ in how the gripper interacts with the environment. While images produce a two-dimensional representation of a three-dimensional scene, in contrast tactile arrays generate a good one-to-one contact representation \cite{navarro2023}. In this way, it is possible to reconstruct the shape of an unknown object in both discontinuous and continuous ways, by sliding while using tactile arrays.
        
        Somatosensory maps inspire tactile servoing controllers \cite{denei2015}. The information is split into normal contact forces that must be exerted on the external body, and contact motion. They are additive contributions to a force-velocity controller that is needed to exert a target force along a trajectory. Somatosensory maps highly affect the performance of the controller, being used for both sensing and planning and allowing an efficient integration of several tactels.

\begin{figure}[t]
            \centering
            \includegraphics[width=.7\linewidth]{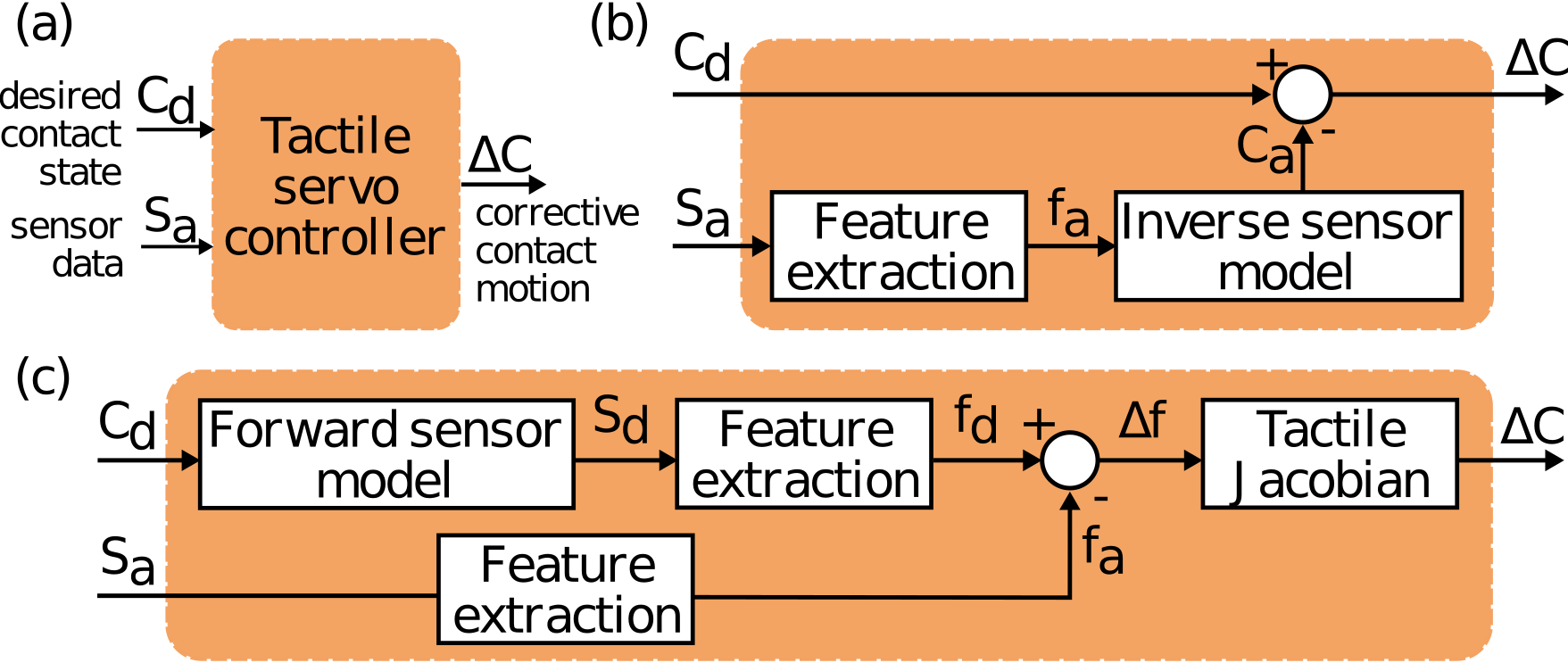}
            \caption{\textbf{(a)} The general tactile servo controller. Starting from the current tactile image $S_a$ to reach a goal contact state $C_d$, the tactile servo controller outputs how much the contact should change $\Delta C$. \textbf{(b)} The \textit{inverse sensor model} maps the features $f_a$ extracted from $S_a$ into the current contact $C_a$; \textbf{(c)} the \textit{forward sensor model} links the desired contact to the expected tactile image $S_d$, whose feature vector $f_d$ is computed from. It is then compared to the current feature $f_a$, and the contact motion is generated by mean of the \textit{Tactile Jacobian} and the residual feature vector $\Delta f$.}
            \label{fig1}
        \end{figure}

        If the task planner gives the robot positioning, tactile images are used to estimate the contact state \cite{son1996b, delgado2016}.  Tactile servo controllers - shown in Figure \ref{fig1} - have been proposed to track tactile evolution \cite{zhang2000}. The \textit{inverse sensor model} maps sensory data to the contact state, which is carefully constrained to avoid ambiguity or degeneration. In the case of sensor multi-modality, the robot can partially interprets contact information with other modalities and simplify the process of exploiting tactile images \cite{liu2017}. Few simple contact states can be estimated from the tactile image, yet highly sensitive to noise. 
        
        On the other hand, the \textit{forward sensor model} maps a given contact state into the expected numerically computed tactile image, and then tactile feature vectors are extracted. The model works in combination with the Tactile Jacobian, a mathematical object that receives as input the residual tactile vector generated by desired and actual tactile images and outputs the amount of correction necessary to enslave the target in the contact space. The Tactile Jacobian method is therefore applicable to a broader range of contact cases than its inverse counterpart. But, as with many numerical solutions, the Tactile Jacobian does not lend itself to a convenient representation since its computational complexity increases exponentially with the dimension of the contact state-space. Conversely, the solution based on the inverse model has been implemented for edge tracking \cite{kappasov2020} to map the tactile feature error to a corrective action.

        Combining tactile servoing with high-level planner - like target alignment - can significantly improve robotic manipulation tasks. For instance, it allows for precise and robust robotic pushing \cite{lloyd2022,ozdamar2024}. The tactile servoing offers real-time adjustments based on tactile feedback, while the high-level planner provides the overarching strategy, allowing for accurate, goal-oriented tactile exploration under various conditions. Additional implementations involve real-time tracking and pose reconstruction \cite{stuber2020}, planar objects shape and configuration estimation \cite{suresh2021}, and smooth spatial tactile servoing by incorporating shear forces \cite{lloyd2024poseshear}.

        \subsection{Discussion}
        Tactile exploration is enabled by capabilities of incremental complexity - as also depicted from Table \ref{tab:summary_table} - from contact making to tactile servoing. Exploration generates an estimate of local and global characteristics, used to build an inner representation of the external world. Such knowledge is furtherly exploited for active exploration and force-velocity control, while keeping stable contacts.
        
        As an alternative to model-based approaches, learning methods \cite{yang2021,kim2019} have been used for gripper-object contact estimations, to link sensors’ output to tactile features (i.e., contact position, force and torque). Despite getting rid of modelling, data-driven approaches have still some working issues, such as training time \cite{kim2019} and data incompleteness \cite{yang2021, cao2023incomplete}. On the other side, the estimation of inter-objects contact distributions \cite{kroemer2018} can be improved by inspecting additional information, such as material properties and contact motion. The application to multi-object and multi-point contacts estimation is still under investigation.
        
        Pure tactile-driven \cite{pestell2019,defarias2021} or multi-sensory fusion \cite{bimbo2015,calandra2018,donato2024multimodal,donato2024xai} methods can be used to interact with unknown objects. Tactile recognition and contact estimation are improved if the number of tactels is increased and if all the sensors are kept in contact with external objects \cite{gil2016}. Tailoring a proper grasp quality metrics is fundamental since it may cause over-adjustments and consequently limit the grasp robustness \cite{pestell2019}. Both spatial and temporal dimensions are fundamental to gather informative tactile signals, but cues on the optimal resolution for both parameters is still being researched, to get a trade-off between task performance and body sensorization.
        
        Contact information can be then exploited for tactile exploration. It can enable surface reconstruction, increasing the throughput \cite{yi2016,saund2021,suresh2022,rustler2022,bimbo2022}. Anyway, it generally assumes the exploration of only fixed and rigid objects. Consequently, the tactile shape exploration of deformable structures and movable parts should be investigated \cite{liao2024expgraph}. It allows pushing active sensing in real case scenarios, since contact forces may cause object displacement and deformation.
        
        Stable contact formation is not merely a static task but also involves dynamic processes such as tactile servoing \cite{kappassov2022}. It is used to guide the end effector across a target surface to evaluate properties like an object's softness \cite{delgado2016}, its geometric shape \cite{cannata2010}, or to navigate across deformable surfaces \cite{kappasov2020}. This exploration usually involves palpation, which can lead to undesirable discontinuous contacts.

        Challenges associated with implementing tactile servoing can be divided into control problems and uncontrolled sensory information. For control problems, issues can arise when the target contact position defined by the controller falls outside the workspace of the finger. Moreover, high contact velocities can alter the sensor response. Uncontrolled sensory information issues are typically tied to the quality of object features and the unpredictability of tactile sensing. Uncontrolled variations can potentially impact the system's stability. Advancements in the implementation of tactile servoing could consider the adoption of distributed tactile sensing, which has been only partially explored in the context of robotic hands. Distributed tactile sensing could offer a more comprehensive and nuanced understanding of the tactile landscape, enhancing the stability of tactile servoing.

    \begin{landscape}  
        \begin{table}[ht]
            \footnotesize
            \centering
            \resizebox{2.2\textwidth}{!}{
            \begin{tabularx}{4\textwidth}{@{}lllllllllll@{}}

                \textbf{Task} &
                  \textbf{Sub-problem} &
                  \textbf{Challenge} &
                  \textbf{Tactile sensor} &
                  \textbf{Multi-modality} &
                  \textbf{Robotic effector} &
                  \textbf{Model/Architecture/Control} &
                  \textbf{Input data} &
                  \textbf{Output} &
                  \textbf{Performance} &
                  \textbf{} \\ \cmidrule{1-10}
                \multirow{6}{*}{Active exploration} &
                  \multirow{3}{*}{Local exploration} &
                  Object property estimation &
                  Tactile array &
                  - &
                  Barret hand &
                  Deep-LSTM network &
                  Tactile images &
                  Sample hardness &
                  99\% of accuracy &
                  \cite{fang2022} \\
                 &
                   &
                  Tactile fusion over space &
                  GelSight &
                  - &
                  Baxter hand &
                  BRISK, RANSAC &
                  Tactile images &
                  Object orientation &
                  95\% of accuracy &
                  \cite{li2014a} \\
                 &
                   &
                  Curiosity-driven exploration &
                  Tactile array &
                  - &
                  Custom finger &
                  Reinforcement Learning &
                  Tactile data &
                  Surface recognition &
                  92\% of accuracy &
                  \cite{pape2012} \\ \cmidrule{2-2}
                 &
                  \multirow{3}{*}{Global exploration} &
                  Shape recognition &
                  Tactile array &
                  Proprioception &
                  Phantom Omni &
                  iCLAP &
                  Tactile images, Arm configuration &
                  Object recognition &
                  95\% of accuracy &
                  \cite{luo2019,gupta2022} \\
                 &
                   &
                  Entropy-driven exploration &
                  Force sensor &
                  - &
                  Allegro Hand &
                  GPIS + Local GP &
                  Contact force and location &
                  Object representation &
                  - &
                  \cite{khadivar2023} \\
                 &
                   &
                  Intrinsic motivation &
                  Force sensor &
                  Proprioception &
                  Robotic arm &
                  Reinforcement Learning &
                  Contact force, Arm configuration &
                  Exploration policy &
                  - &
                  \cite{vulin2021} \\ \cmidrule{1-10}
                \multirow{2}{*}{Tactile servoing} &
                  Model-less control &
                  \begin{tabular}[c]{@{}c@{}}Combine tactile\\ array information\end{tabular} &
                  Tekscan sensor &
                  Proprioception &
                  Shadow hand &
                  PID &
                  \begin{tabular}[c]{@{}c@{}}Contact force,\\ Finger configuration\end{tabular} &
                  Deformability degree &
                  - &
                  \cite{delgado2016} \\ \cmidrule{2-2}
                 &
                  \begin{tabular}[c]{@{}c@{}}Sensor model\\ implementation\end{tabular} &
                  \begin{tabular}[c]{@{}c@{}}Tactile information\\ to contact estimation\end{tabular} &
                  Tactile array &
                  Proprioception &
                  - &
                  Contact model &
                  \begin{tabular}[c]{@{}c@{}}Contact force,\\ Finger configuration\end{tabular} &
                  Surface reconstruction &
                  - &
                  \cite{zhang2000} \\ \cmidrule{1-10}
                \multirow{3}{*}{Pre-shaping} &
                  Tactile impairment &
                  \begin{tabular}[c]{@{}c@{}}Dependence of touch\\ on effector sinergies\end{tabular} &
                  ThimbleSense &
                  Proprioception &
                  Human hand &
                  Kinematic model &
                  Contact force, Hand configuration &
                  Succesful pre-shaping &
                  - &
                  \cite{dellasantina2017} \\ \cmidrule{2-2}
                 &
                  \multirow{2}{*}{\begin{tabular}[c]{@{}c@{}}Multi-modal\\ preshaping\end{tabular}} &
                  \multirow{2}{*}{\begin{tabular}[c]{@{}c@{}}Tactile prediction\\ for pre-shaping\end{tabular}} &
                  - &
                  Vision &
                  ARTS humanoid &
                  Predictive architecture &
                  RGB images &
                  Tactile image prediction &
                  70\% of success &
                  \cite{laschi2008} \\
                 &
                   &
                   &
                  - &
                  Proximity &
                  DH-2 hand &
                  Reactive control &
                  Proximity information &
                  Pre-shaping pose &
                  - &
                  \cite{koyama2013} \\ \cmidrule{1-10}
                \multirow{5}{*}{Static holding} &
                  \begin{tabular}[l]{l}Dealing with\\ unknown objects\end{tabular} &
                  Fragile object holding &
                  FBG &
                  - &
                  Cam-Hand &
                  PID &
                  Tactile information &
                  Finger velocity &
                  99\% of accuracy &
                  \cite{massari2019} \\ \cmidrule{2-2}
                 &
                  \multirow{2}{*}{Optimal holding} &
                  \multirow{2}{*}{Optimal hand closure} &
                  Inertial, pressure &
                  - &
                  \multirow{2}{*}{Two-fingers gripper} &
                  \multirow{2}{*}{Fuzzy logic} &
                  Multi-modal tactile information &
                  \multirow{2}{*}{Joints' torque} &
                  - &
                  \cite{dafonseca2019} \\
                 &
                   &
                   &
                  Pressure sensor &
                  - &
                   &
                   &
                  Pressure &
                   &
                  - &
                  \cite{kim2020} \\ \cmidrule{2-2}
                 &
                  \multirow{2}{*}{Learn from experience} &
                  Manage grasp shape memory &
                  Tactile array &
                  Proprioception &
                  Barret hand &
                  SVM &
                  \begin{tabular}[c]{@{}c@{}}Experience database,\\ Tactile images\end{tabular} &
                  Updated database &
                  - &
                  \cite{zhang2015b} \\
                 &
                   &
                  Hold deformable objects &
                  Magnetostrictive sensor &
                  - &
                  Two-fingers gripper &
                  Extreme learning machine &
                  Tactile data &
                  Stiffness estimation &
                  95\% of accuracy &
                  \cite{zhang2018} \\ \cmidrule{1-10}
                \multirow{8}{*}{Grasping} &
                  \multirow{2}{*}{\begin{tabular}[c]{@{}c@{}}Incipient slip\\ suppression\end{tabular}} &
                  Slippage of unknown objects &
                  Tactile array &
                  - &
                  Two-fingers gripper &
                  Fuzzy logic &
                  \begin{tabular}[c]{@{}c@{}}Tactile features\\ (velocity, acceleration, etc.)\end{tabular} &
                  Fingertip velocity &
                  - &
                  \cite{glossas2001} \\
                 &
                   &
                  Sampling time influence &
                  Pressure sensor &
                  - &
                   &
                  Ensemble of Bagged Trees &
                  Pressure &
                  Slip detection &
                  80\% of accuracy &
                  \cite{levins2020} \\ \cmidrule{2-2}
                 &
                  \multirow{3}{*}{\begin{tabular}[c]{@{}c@{}}Friction-informed\\ control\end{tabular}} &
                  Constant friction reference &
                  Force sensor &
                  - &
                  Salisbury hand &
                  Forces friction cone, PID &
                  Contact force and location &
                  Stable grasp &
                  - &
                  \cite{bicchi1989} \\
                 &
                   &
                  Online friction estimation &
                  BioTac &
                  - &
                  Custom finger &
                  Coulomb friction law &
                  Normal force &
                  Force, Slip Classification &
                  80\% of accuracy &
                  \cite{su2015} \\
                 &
                   &
                  \begin{tabular}[c]{@{}c@{}}Mechanical friction\\ augmentation\end{tabular} &
                  Optoforce &
                  Vision &
                  Parallel gripper &
                  Soft finger contact model &
                  Normal force, RGB images &
                  Normal force &
                  $< 10^{-3}$ [N] error &
                  \cite{vina2016} \\ \cmidrule{2-2}
                 &
                  \begin{tabular}[c]{@{}c@{}}Robustness\\ enhancement\end{tabular} &
                  Finger repositioning &
                  Capacitive array &
                  - &
                  iCub hand &
                  \begin{tabular}[c]{@{}c@{}}Hierarchical controller,\\ Gaussian mixture regression\end{tabular} &
                  Fingertips' force &
                  Grip strength &
                  - &
                  \cite{regoli2016} \\ \cmidrule{2-2}
                 &
                  \multirow{2}{*}{Stability} &
                  Multi-modal force regulation &
                  Pressure sensor, IMU &
                  Vision &
                  Soft gripper &
                  LSTM &
                  Pressure, RGB images &
                  Stable grasp &
                  66\% of accuracy &
                  \cite{zimmer2019} \\
                 &
                   &
                  \begin{tabular}[c]{@{}c@{}}Friction-dependent\\ force regulation\end{tabular} &
                  3-axes tactile sensor &
                  Proprioception &
                  Multi-fingered hand &
                  Impedance control &
                  Force, finger joints &
                  Stable grasp &
                  - &
                  \cite{zhang2015a} \\ \cmidrule{1-10}
                Releasing &
                  \begin{tabular}[c]{@{}c@{}}Stable objects\\ placement\end{tabular} &
                  \begin{tabular}[c]{@{}c@{}}Discriminate among\\ pulling forces and perturbations\end{tabular} &
                  BioTac &
                  - &
                  Shadow hand &
                  Bayesian estimation &
                  Load force &
                  Finger force &
                  - &
                  \cite{eguliuz2019} \\ \cmidrule{1-10}
                \multirow{8}{*}{Manipulation} &
                  \multirow{2}{*}{Phases identification} &
                  Control points identification &
                  Piezoresistive sensor &
                  \begin{tabular}[c]{@{}c@{}}Vision,\\ Proprioception\end{tabular} &
                  KUKA arm &
                  Impedance control &
                  \begin{tabular}[c]{@{}c@{}}Contact force and location,\\ Joint angles\end{tabular} &
                  Finger position &
                  - &
                  \cite{li2013} \\
                 &
                   &
                  Phase coordination &
                  Camera-based sensor &
                  \begin{tabular}[c]{@{}c@{}}Vision,\\ Proprioception\end{tabular} &
                  Franka Emika panda &
                  \begin{tabular}[c]{@{}c@{}}Dynamic Motor Primitives,\\ Reinforcement Learning\end{tabular} &
                  \begin{tabular}[c]{@{}c@{}}Contact force, Joint angles,\\ RGB images\end{tabular} &
                  Manipulation policy &
                  96\% of success &
                  \cite{narita2021} \\ \cmidrule{2-2}
                 &
                  \multirow{2}{*}{Action planning} &
                  \begin{tabular}[c]{@{}c@{}}Learning from\\ multiple demonstrations\end{tabular} &
                  \begin{tabular}[c]{@{}c@{}}Depth camera-based\\ sensor\end{tabular} &
                  - &
                  KUKA LBR &
                  \begin{tabular}[c]{@{}c@{}}Hidden Markov model,\\ Gaussian mixture regression\end{tabular} &
                  Tactile images &
                  End-effector velocity &
                  - &
                  \cite{huang2020} \\
                 &
                   &
                  \begin{tabular}[c]{@{}c@{}}Online sequential\\ decision-making\end{tabular} &
                  6-axes force sensor &
                  - &
                  KUKA, PR2 &
                  Markov decision process &
                  Tactile information &
                  Robot action &
                  - &
                  \cite{vien2015} \\ \cmidrule{2-2}
                 &
                  \multirow{4}{*}{In-hand manipulation} &
                  Controlled slippage &
                  \begin{tabular}[c]{@{}c@{}}Optical 3-axes\\ force sensor\end{tabular} &
                  - &
                  Parallel gripper &
                  Attitude control &
                  Contact force &
                  \begin{tabular}[c]{@{}c@{}}Loose grip state,\\ Hole detection\end{tabular} &
                  - &
                  \cite{ohka2018} \\
                 &
                   &
                  \begin{tabular}[c]{@{}c@{}}Change grasp configuration\\ while holding\end{tabular} &
                  Force sensor &
                  Proprioception &
                  Two-fingers gripper &
                  Rolling contact model &
                  Contact force and location &
                  Next contact state &
                  - &
                  \cite{maekawa1995} \\
                 &
                   &
                  Imitation learning &
                  Contact sensor &
                  \begin{tabular}[c]{@{}c@{}}Vision,\\ Proprioception\end{tabular} &
                  Shadow hand &
                  Markov decision process &
                  \begin{tabular}[c]{@{}c@{}}Force, joints values,\\ RGB images\end{tabular} &
                  Manipulation policy &
                  90\% of success &
                  \cite{jain2019} \\
                 &
                   &
                  \begin{tabular}[c]{@{}c@{}}Interaction with\\ the environment\end{tabular} &
                  6-axes force sensor &
                  \begin{tabular}[c]{@{}c@{}}Vision,\\ Proprioception\end{tabular} &
                  KUKA arm &
                  Variational AutoEncoder &
                  \begin{tabular}[c]{@{}c@{}}Force data, Robot state,\\ RGB image\end{tabular} &
                  Manipulation policy &
                  - &
                  \cite{lee2020} \\ 
                  \cmidrule{1-10}
            \end{tabularx}}
            \caption{Summary of main tasks in tactile robotics control, highlighting challenges and proposed solutions in the state-of-the-art} \label{tab:summary_table}
        \end{table}
    \end{landscape}

    \color{black}
    \section{Closed-loop Contact Grasping}
        \label{section:grasping}
        A constant gripping force may not ensure a stable holding, thus a slippage. Therefore, feedback strategies become essential. Tactile sensing provides valuable information about the grasping state to achieve damage-free and gentle grasps. End-effector positioning and grasping can be evaluated based on four fundamental properties: disturbance resistance, dexterity, equilibrium, and stability \cite{roa2014}.  In selecting the optimal grasp, various quality measures for contact estimation and hand configuration are considered. Tactile sensing plays a fundamental role in the assessment, except for dexterity, which primarily depends on finger kinematics.

        \subsection{Tactile sense informs pre-shaping}
        The impact of tactile impairment on hand synergies has been analyzed \cite{dellasantina2017}. The impairment does not significantly affect overall movements, but it does increase the time required for pre-shaping and contact-making phases. Additionally, impaired individuals exhibit larger forces compared to unimpaired ones. Inertial sensing in soft robotic hands minimizes the applied force on external objects.
        
        However, pre-shaping is not solely reliant on tactile information, as sensors cannot gather data until contact is established. Visual and proximity information plays a crucial role. Laschi et al. \cite{laschi2008} propose a learning strategy that incorporates vision-based tactile prediction, enabling a tactile information-dependent pre-shaping. Alternatively, Koyama et al. \cite{koyama2013} utilize a network of proximity sensors to perform online preliminary motions before grasping, further adjusted after the contact has been made.
        
        \subsection{Object holding}
        Effective object-holding control allows for stable grasping, precise manipulation, and safe transportation of objects. It enables tasks such as picking up, carrying, releasing, and positioning objects with accuracy and control.
        
        Similar to tactile exploration, tactile maps are put into play for grasp planning. The discepancy between real-time tactile images and the expected ones acts as an error to minimize to enhance the control of the grasp \cite{delgado2017a}. More precisely, the local error is the control parameter for the robotic finger's movement. Forces are mapped onto natural movements and then translated into joint velocity. Alternatively, tactile sensing helps maintaining stable pinching, even in the face of tactile measurement errors \cite{psomopoulou2021}. Despite its general-purpose suitability, indirect detection and remedial action have limited reactivity, potentially leading to object dropping.
        
        Grasping novel objects can pose challenges without prior experience. A tactile-based approach to object classification and holding is outlined in \cite{massari2019}. Two types of controllers, static and dynamic, guide the sensorized gripper from a free to a grasping position. The static controller uses a fixed voltage input for constant speed movement, while the dynamic adjusts speed based on sensor feedback to changes in size.


        Alternatively, optimizing grasp metrics can facilitate diverse grasping strategies, like in tactile-based fuzzy control. In \cite{kim2020}, a fuzzy controller employs contact pressure and its derivative to determine the optimal grasping configuration and forces. Conversely, finger-level fuzzy feedback controllers assess pressure status and grasp stability based on pressure and micro-vibration readings \cite{dafonseca2019}. The output of this initial stage is input to a subsequent layer, which produces finger action commands (or motor velocities) to ensure stable grasping. Optimization can also be exploited to find the best contact position for grasping and to exert forces high enough to slide objects on surfaces up to a goal position \cite{thompson2022}. The controller always involves a planning strategy that exploits the robot model.
        
        From a different perspective, learning and storing information while exploring unseen objects can generate better outcomes \cite{kuwashawa2024cl}. A continuous comparison with an experience database, which stores tactile and proprioceptive information of the grasp, allows to assess if a new grasp has been experienced and is stable, and then performed to grip the object \cite{zhang2015b}; otherwise, the experience is stored and a joint-level torque adjustment is performed. Variants of this solution exploit also shape similarities between different objects to drive stable grasping \cite{dang2014} or interleaved local exploration and grasp adaptation, using a virtual frame to estimate the object pose \cite{li2014b}. The impedance controller is not focused on the object shape but on contact points: the aim is to find a proper set of fingertips positioning from the virtual frame and stiffness parameters to assess a stable and adaptable grasp.
        
        Learning from previous interactions could increase the grasping success especially if in contact with deformable bodies. Such a clue is evident in \cite{zhang2018}, in which a robot learns how to perform a haptic intensive manipulation task through Contextual Multi-Armed Bandits, while decreasing the training time and increasing the cumulative rewards over a finite period. The model-free classifier learns the nonlinear mapping between finger pad deformations and the relative location of a deformable contour along the length of the finger pad. It is important to limit the dimensionality of the state-action space to make the exploration problem tractable, especially given the high cost of gathering tactile sensor data. Tactile signals impact on the reward and the policy state vector to identify tactile sensing needs in reinforcement learning grasping algorithms. Indeed, Koenig et al. \cite{koenig2022} show how much tactile information is essential to exploit grasp metrics in the reward function, while decreasing the sensor resolution does not considerably impact on the algorithm performance, with implications on grippers design and training. 
        
        At the end of the grasp, the robotic gripper leaves the object, and tactile sense enables the discrimination between the two phases. A method to recognize between object pulling forces and random perturbations during object handover is implemented \cite{eguliuz2019}. Robotic fingers are not controlled in position but in force, to maintain the same contact force direction and magnitude. Eguiliuz et al. suggest that such control implements compliance in contact force.

        \subsection{Grip control for slip suppression}
        Slippage sensing is essential for grasping with minimal force and preventing objects from falling \cite{wang2019a}. Avoiding slippage is a strategy to improve grip stability by regulating gripping forces without squeezing the objects. The most trivial solution is to reactively increase contact forces in the event of slippage \cite{li2018a}, regardless of whether damage occurs. 

        Slippage detection remains a challenge since it may result from either the slipperiness of the object or an external disturbance \cite{deng2020}. Concerning contact mechanics, the onset and direction of slip can be inferred by changes in shape and size of the area that sticks to the surface \cite{delhaye2014}. The choice for proper slippage contact sensors is not random: indeed,  all human hand mechanoreceptors, but slowly-adapting receptors are activated when the target does not slip \cite{srinivasan1990}. Micro- and macro-slip phenomena can be sensed using vibration, pressure, and vision sensors. For instance, \cite{tada2012} correlates the output of receptors with visual information to sense slippage through pick-up experiences, and controlling the distance between fingers based on a slip detection signal. Features can also be extracted from motion, but are constrained by the mechanical filtering of elastic covers \cite{shimojo1997} since sensors with high rigidity are not suitable for slippage detection. 

        Both frequency analysis and motion estimation techniques can be used for slippage suppression \cite{ji2019}. When the relative speed of fingers and objects is relatively low, either the slippage is an intermittent vibration in the gripping force detected by the high-frequency signal it generates, or the centre of mass tracking through convolutional methods provides information about the slip distance. Both are employed to update the grip force at each timestep by a factor $\alpha\beta$, where $\alpha$ is based on the object stiffness and $\beta$ is the slip signal. A high-stiffness object will have a bigger $\alpha$ to quickly eliminate slippage; a low-stiffness object will have a smaller $\alpha$ to avoid distortion of the object by gently increasing the grip force. Alternatively, the BioTac sensor has been used for event-based slippage prediction and grip stabilization \cite{veiga2018}, with some insights on feature selection due to high-dimensional tactile information. If slippage is detected, the controller increases the normal fingertip velocity to adjust the normal force. Similarly, object slippage can be detected with pressure sensors only \cite{begalinova2020}.

        \begin{figure}[b]
            \centering
            \includegraphics[width=.7\linewidth]{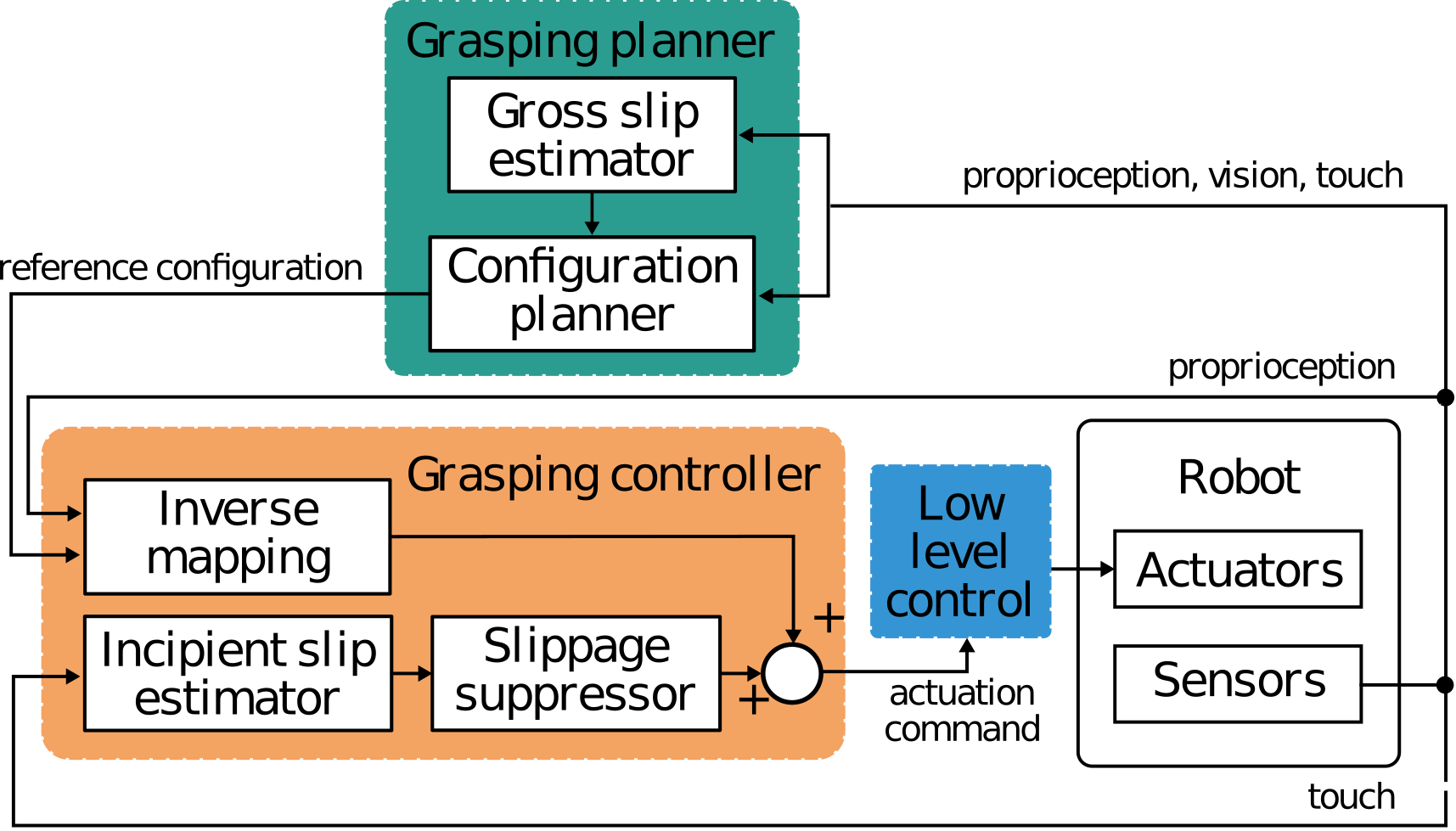}
            \caption{This controller aims to prevent incipient slippage, and adjusts the grasping configuration if gross slippage occurs. Tactile features provide information for slippage detection, allowing the control of fingertip velocity or contact forces for its suppression. At a higher level, the grasping planner estimates and considers gross slippage, generating a reference configuration for the gripper to grasp the target.}
            \label{fig:slippage_controller}
        \end{figure}
        
        Slippage can be observed in two distinct phases. The \textit{incipient slip} occurs when the object is on the verge of falling. During this phase, tangential and normal forces provide precise information on the surface friction coefficient, aiding the robot's controller in preventing subsequent major slips by applying suitable contact forces \cite{chen2018}. For instance, the predictive model by Petchartee et al. \cite{petchartee2008} can detect incipient slip between three and six stages before the \textit{gross slip} happens. Figure \ref{fig:slippage_controller} illustrates a comprehensive control architecture to address both challenges.
        
        Tactile feedback can modulate finger velocity to prevent slippage without knowing the geometrical or physical characteristics of the grasped object. The grip force update rule to counteract slippage in \cite{veiga2020b} permits to act without any prior information about the object. Alternatively, slip suppression could be driven by more intelligent fuzzy systems, by either using euromorphic information from a camera-based tactile sensor to compute corrective actions \cite{muthusamy2020} or implementing a dynamic slip control \cite{glossas2001}, where the force must be kept low without overshooting. The rules set is created by investigating the human hand, and the fuzzy system permits to achieve a gripper behaviour close to the natural counterpart. 

        If no manipulation template is available, data-driven approaches might be considered. A pneumatic tactile sensor provides multiple abilities such as contact force computation and slip detection through machine learning \cite{levins2020}. Data have been collected while both speed and force between sensor and object change. A shift to the frequency domain is needed, and the top five dominating frequencies are used for classification. Shaw et al. built a robust time-discrete controller for unknown objects tactile-driven manipulation \cite{shaw2020}. Semiglobal practical asymptotic stability of the closed-loop system is guaranteed. The tuning parameter of the control system is the sampling time: as it decreases, the system asymptotically converges to a smaller bound around the reference.

        \subsection{Friction-informed grasping control}
        While primary controllers often overlook friction forces, some models assume a known constant friction coefficient to mitigate the impact of normal finger force. Enhancing grasp performance in unknown environments could be achieved through online friction estimation. However, not all manipulation models can offer closed-form solutions, thereby necessitating the use of data-driven methods.

        
        If the friction coefficient $\mu$ is assumed to be known, the online tactile information can be used to assess stable grasps \cite{bicchi1989}. Tactile sensors output tangential $f_t$ and normal $f_n$ forces, and torque $m$ generated by friction forces. The slipping-safe region for a grasp is generated by $\mu m + K f_t \le K \mu f_n$, where $K$ is a function of the parameters of the grasped object and $\mu$ is the dry friction coefficient. The target normal force that might prevent slippage and eventual damages should be as close as possible to its desirable value $F_{norm}$ in the absence of disturbances, and the distance from the slippage region boundary should be as large as possible. This strategy is used for grasp force optimization and control, consisting of real-time feedback from contact sensors on fingertips.

        Online estimation of the friction coefficient could be a safer solution to avoid slippage, instead of conservative \cite{wettels2009,romano2011} or offline estimation \cite{demaria2015}. The force-position controller in \cite{su2015} allows robotic fingers to interact with a target object and adapt the normal force at the fingertip level to obtain a gentle grasp. An initial guess of the friction coefficient is provided and updated if the normal force leads to a slip.


        \subsection{To improve grasp robustness and stability}        
        

        Using near-field sensor data enhances grasp robustness \cite{lynch2021}, allowing the hand to reposition and keep objects centred during a grasp \cite{dollar2010}. This approach expands the gripper's operating area, hence enhancing the grasp stability and contact force balance \cite{pan2018}. A hierarchical grasp controller, on the other hand, can adapt grip strength and stability against external disturbances \cite{regoli2016}. The low-level controller stabilizes the grasp with given grip strength, while the high-level controller modifies the hand configuration and enables re-grasping for optimal manipulation poses.

        \begin{figure}[t]
            \centering
            \includegraphics[width=.7\linewidth]{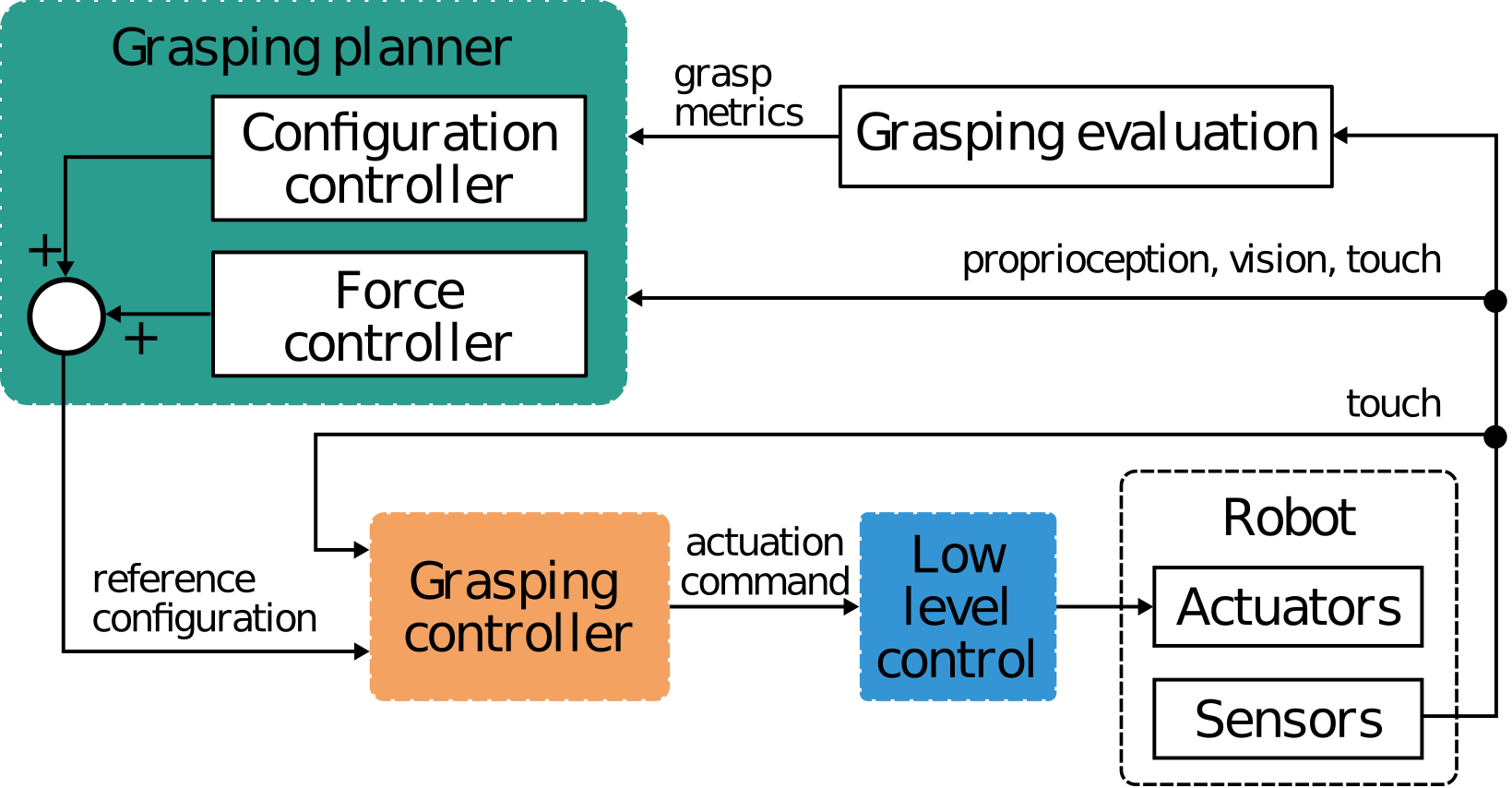}
            \caption{Tactile features help assessing grasping quality metrics. The grasp planner can leverage them to generate more stable and robust configurations.}
            \label{fig:grasping_contoller}
        \end{figure}
        
        As from Figure \ref{fig:grasping_contoller}, modulating force during early grasping is crucial to stabilize the object. When vision is available, the properties of the target object can be recognized in advance to anticipate the correct grip force. For instance, the grip force could be adequately scaled to the object weight \cite{hermsdorfer2011} and the fingertip force to object shape and size  \cite{flanagan2002}. Zimmer et al. identified a LSTM as the most accurate multi-modal estimator to assess grasp success and stability \cite{zimmer2019}. It emphasizes the importance of the temporal dimension: the inertial data primarily captures macro gripper movements, while pressure sensors provide information on smaller movements. 

        Haptic information about an object texture significantly influences grip force modulation before lift-off \cite{tiest2019}. The fingertip tactile signal modulates grasping force; slippage detection triggers an adjustment in force to maintain the total force vector within the friction cone \cite{maekawa1997}. An alternative solution limits each fingertip force within the friction cone, preventing undesired slippage or detachment \cite{zhang2015a}.

        Grasp stability is crucial to both holding an object steady and transitioning it across grip configurations. According to behavioural studies, human grasp control is modular and largely independent per finger due to Independent Finger Control, countering potential slips \cite{edin1992}. This aligns with human motor control observations where the force applied by each digit barely exceeds the minimum to prevent slippage.
        

        \subsection{Discussion}
        Tactile signals provide valuable spatial and temporal information about contact states. However, their high dimensionality poses a challenge, and finding a way to synthesize their content without losing information is crucial \cite{kim2020,dafonseca2019}. This is particularly important for controllers handling real-world tasks, as responsiveness is a key factor for success. It contributes to the training of data-driven algorithms, which often face the problem of exponentially increasing dimensionality in the state-action space, making it intractable \cite{koenig2022}.
        
        The sense of touch is inactive without interaction, but its absence can actually hinder the early, crucial stages of contact to achieve a stable and reliable grasp \cite{dellasantina2017}. Recognizing the importance of tactile feedback, researchers have explored methods for predicting the future tactile response while the robot is approaching an object, to optimize the pre-grasping configuration \cite{laschi2008}. It questions on the role of haptic memory in interaction tasks for both recognition and control \cite{bekiroglu2016}.
        
        Grasp controllers face the challenge of meeting three crucial requirements that impose limitations on the grasp space and demand intricate and sophisticated control strategies. These requirements are driven by challenges reported in Table \ref{tab:summary_table}: (i) ensuring that the physical and geometric properties of the grasped object are not altered, (ii) mitigating the risk of undesired object slippage, and (iii) effectively managing the distribution and balance of contact forces to achieve grasp stability and robustness. While the first requirement primarily influences contact positioning, the latter two constrain the range within which the contact force can vary. 
        
        Existing grasping strategies can be categorized into two approaches: one that involves estimating a model based on gathered information and comparing it to previously encountered objects \cite{zhang2015a,dang2014,zhang2018}, and another that aims to make the controller independent of the physical model and focus solely on the contact features \cite{massari2019,hughes2020}. Consequently, the controller adapts its behavior according to the specific morphology of the gripper, including factors such as contact points and finger morphology, as well as the tactile information available to guide the grasping phase.
        
        To fulfill the first two requirements, the controller needs to exert a grip force that prevents damage to the object while also preventing slippage. The assessment of potential damage can be achieved through the analysis of raw tactile signals, while estimating slippage and friction requires establishing a minimum grip force threshold. Detecting slippage should occur in the early stages, before significant movement of the object is observed \cite{petchartee2008,chen2021,cui2023}. This constraint has sparked discussions on how tactile data should be recorded and processed, with temporal considerations indicating the need for the shortest sampling time required to detect significant frequency features \cite{levins2020,shaw2020}. Although early controllers did not consider friction at all, the first step towards addressing it was through offline estimation using known constant friction coefficients \cite{bicchi1989}. However, physical objects possess non-predictable friction coefficients across their surfaces, suggesting that online estimation \cite{su2015} is necessary to generalize to unknown features.
        
        The controller utilizes the available information to directly modulate the normal force exerted by the fingertip on the grasped object through force control \cite{petchartee2008,watanabe2007,lu2024dexitac}, or indirectly through position or velocity control \cite{veiga2018,dafonseca2019}. This approach can be applied to both rigid or deformable objects, as well as soft grippers. When the friction coefficient is known, it can be utilized to determine the minimum gripping force \cite{su2015} that prevents slipping without causing deformation of the object. Several tools have been mentioned for this purpose, including the friction cone limit \cite{maekawa1997}, independent finger control \cite{edin1992}, and stick ratio \cite{watanabe2007}.


        
        
       

    \color{black}
    \section{Tactile Dexterous Manipulation}
        \label{section:manipulation}
        Tactile-driven robotic handling extends beyond static object grasping and requires adaptation to highly unpredictable situations. The richness of tactile signals facilitates in-hand object re-configuration tasks, thereby enhancing the overall task performance. 

        \subsection{Identifying manipulation phases}
\begin{figure}[b]
            \centering
            \includegraphics[width=.7\linewidth]{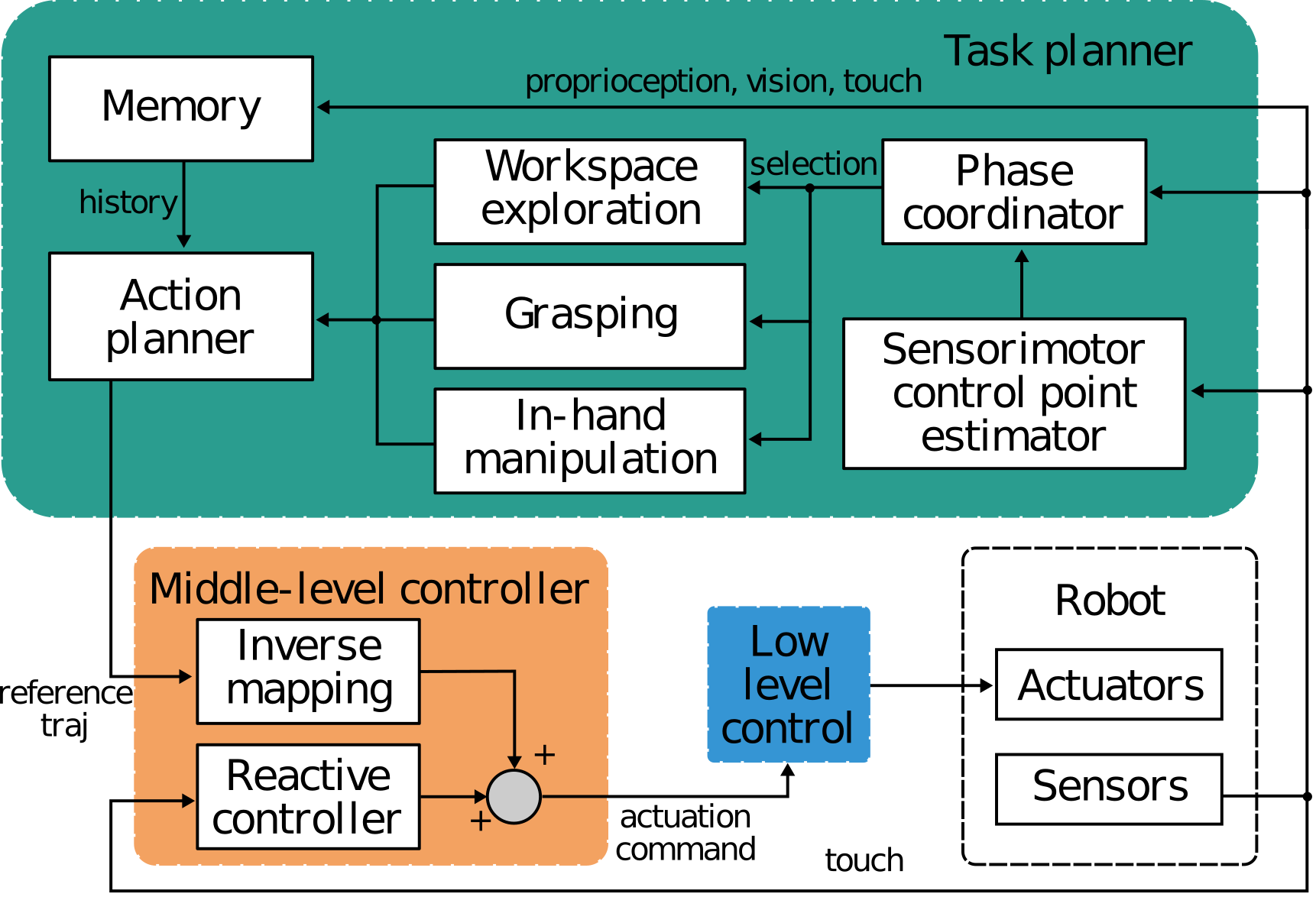}
            \caption{Task phase coordination is facilitated by sensorimotor control point estimation. Upon identifying each task phase, the stage controller plans specific manipulation actions. The haptic memory is harnessed by the action planner to craft predictive or time-dependent strategies. Concurrently, the reactive controller employs sensing to adjust control actions, allowing for real-time adaptation.}
            \label{fig:manipulation_contoller}
        \end{figure}

        Manipulation tasks, like holding and moving objects, entail a sequence of action stages demarcated by mechanical events, like contact initiation and termination. These events, or \textit{control points}, generate distinct sensory signals that provide critical information on the progression and interim goals of the task, thereby playing a pivotal role for control \cite{flanagan2006}. Enhanced discrimination between these stages could be achieved through the use of a skin acceleration sensor, which captures tactile information to enable smoother transitions between stages. Figure \ref{fig:manipulation_contoller} shows how control points can be used to coordinate phase selection and implement contact-dependent strategies.

        In the realm of object manipulation, tactile feedback has shown potential for facilitating smooth and adaptable control. Li et al. \cite{li2013} designed a controller to facilitate in-hand manipulation and mimic human-like manipulation, providing agile responses to unexpected events like slippage. Anyway, planning and control phases are intertwined rather than sequential. This involves a two-step process starting with local object relocation where the object is slightly moved using a single finger while other fingers maintain a stable grasp. This is followed by a global regrasping phase to rearrange the hand configuration for continued local movement. 
        Kim et al. \cite{kim2011} proposed an alternative force/position controller that employs tactile data via a precision motion controller.
        
        When examining grasping and manipulation tasks, some actions like reaching and adjusting are common, while others are specific to the task at hand. Narita et al. \cite{narita2021} designed a \ac{RL} algorithm that decomposes complex tasks into \acp{DMP}. This strategy learns specific actions for each task while recurring actions are learned once and applied across different tasks. A blending policy is later implemented to achieve the task goal based on tactile feedback. Additionally, a hierarchical structure may provide an effective approach for handling complex tasks that necessitate prioritized multi-modal sensing.
                
        \subsection{Planning for interaction}
        Adaptive mechanisms need to be incorporated into the control algorithm to deal with partially known environments. Tactile sensors have proven to be a successful feedback for on-board adaptation, allowing robots to handle unforeseen events during execution without sacrificing system flexibility.

        Camera-based tactile sensors have gained popularity in recent years for manipulation tasks \cite{li2024camera}. They provide rich information that can be analyzed using \ac{DNN}. For example, the OmniTact sensor \cite{ebert2023} is utilized for angle of contact estimation and peg insertion tasks. The sensor data is processed by a deep network to extract 2D position and intensity information. Another example is the DIGIT system \cite{lambeta2020}, which enables in-hand manipulation of marbles. It employs a two-stage learning system where an autoencoder processes the tactile image to obtain state information, which is then used with actions to train a forward dynamics model. \ac{MPC} is utilized to generate optimal action sequences for achieving task goals. A Video Prediction Architecture with long skip connections on the GelSight sensor predicts the next status of the tactile image based on the current input \cite{tian2019}. A \ac{MPC} algorithm plans the best actions over a predictive horizon for marble manipulation. Alternatively, \ac{LfD} is leveraged to perform manipulation tasks with a soft tactile sensor \cite{huang2020}. The robot learns from few demonstrations to follow pipe profiles with various orientations and positions in space, as well as to perform movements with simple objects. 

        \acp{POMDP} have demonstrated their effectiveness in tactile-driven localization in robotics \cite{vien2015}. \acp{POMDP} provide a robust hierarchical framework for online sequential decision-making. The state-space of a \ac{POMDP} consists of hidden states that can only be inferred through observations, such as contact states. By employing sequential submodular optimization techniques, subtasks composed of primitives can be designed to satisfy adaptive submodular optimization criteria and disambiguate uncertainty factors.
    
        \subsection{In-hand object manipulation}
        Dexterous manipulation involves altering object configurations without placing them, relying on precise control and coordination of multiple fingers \cite{ozawa2017}. Central to this capability is the use of tactile signals, which help estimate an object's current pose and provide feedback for sliding or rolling motions within the grasp space.

        Building on this, Chen et al. \cite{chen2021} leveraged convolutional networks to identify objects and measure sliding velocities. Their system integrates tactile and visual data during training, but it relies solely on onboard SynTouch BioTac sensors for feedback during operation. The estimator maps tactile signals to sliding velocities, overcoming the complexities of physical interactions. Beyond sliding control, the architecture detects slippage and dynamically adjusts gripping forces, enhancing overall manipulation reliability.
        
        A significant advancement in dexterous manipulation is controlled grip release, which adds additional \acp{DoF} for manipulating objects \cite{ohka2018}. These added \acp{DoF} enable finer control for tasks such as inducing relative motion between the gripper and object, reconfiguring grasps while holding an object, or improving manipulability through actuator-based \acp{DoF}. For example, Maekawa et al. \cite{maekawa1995} demonstrated the use of kinematic analysis and tactile feedback to guide objects along desired paths, while Reynaerts et al. \cite{reynaerts1994} allowed cylindrical objects to roll over a thumb surface during whole-finger manipulation.
        
        Tactile information is also vital for enhancing object stiffness control in multi-fingered robotic hands \cite{son1996a}. For in-hand object rolling, Ward-Cherrier et al. \cite{wardcherrier2016} proposed a model-free TacThumb-based controller that follows online trajectories without requiring a kinematic hand model. This approach generalizes well to various tactile tasks, expanding its applicability. Similarly, active shape-changing—achieved by inflating pneumatic chambers—enables translational or rotational object movements. Fingertip tactile pressure sensors close the control loop, allowing precise adjustment of object configurations during manipulation steps \cite{he2020}.
        
        Grasp reliability further improves when objects are held closer to their \ac{CoM}, reducing torque and stabilizing the interaction. Li et al. \cite{li2019a} combined external cameras and GelSight sensors to detect object rotation, using early fusion techniques to merge tactile images with other sensor data. This approach enhanced interaction stability and slippage prevention, even for slippery or smooth objects where visual cues are more informative than tactile ones \cite{li2018b}.
        
        Despite extensive research into grasping shapes \cite{miller2004, lenz2015}, challenges persist in manipulating objects with variable stiffness \cite{matulevich2013}. Addressing these challenges, Fukui et al. \cite{fukui2012} introduced a master-slave controller where index and middle fingers are position-controlled, and the thumb is force-controlled, with movements coordinated through tactile feedback. This approach supports both exploration and manipulation. Similarly, Veiga et al. \cite{veiga2020a} proposed a hierarchical framework that combines low-level stabilizing controllers with high-level policies. This framework enhances learning efficiency, real-world transferability, and overall performance in dexterous manipulation.
        
        Deep learning further expands the potential of multi-modal sensory fusion for improved interaction with the environment \cite{jin2024visuotactile,mao2024}. Lee et al. \cite{lee2020} demonstrated this in peg insertion tasks using visual-tactile fusion, while Cui et al. \cite{cui2020} employed a 3D neural network to merge sensory inputs. By optimizing temporal parameters and integrating multiple sensory sources, their approach achieved high accuracy in assessing grasp states for deformable objects. Similarly, tool manipulation on deformable objects using tactile feedback has been realized \cite{zhang2023}.
        
        \subsection{Discussion}
        Object manipulation hinges significantly on the gripper morphology \cite{ozawa2017}, including the number of contact points and \acp{DoF} available to the end-effector. This complexity deepens when considering in-hand object re-configuration, where the object is already within the grasp.

        One crucial challenge in object manipulation is managing slippage. Though detrimental to a secure grasp, strategic slippage can introduce an additional degree of freedom to the system \cite{maekawa1995}. By astutely harnessing gravity, it is possible to reduce grasping force to induce controlled slippage \cite{ohka2018,xu2023,toskov2024,gloumakov2024}, opening new avenues for altering the object pose.
        
        Active relocation strategies, which entail deliberate motion of the grippe \acp{DoF}, can enhance the manipulation process. By augmenting the system \acp{DoF}, it allows for more dexterous manipulation \cite{reynaerts1994,wardcherrier2016}. Two-handed manipulators exemplify this, their advanced cooperative capabilities facilitating sophisticated control and coordination \cite{ishi2006,ohka2012}.
        
        Incorporating additional sensory inputs like proprioception or vision can provide a richer understanding of the manipulation process \cite{li2019a}. Indeed, Table \ref{tab:summary_table} shows a higher need for multi-modality than previous discussed tasks. The integration of multi-modal signals offers a comprehensive view of the manipulation phenomenon, facilitating the identification of different manipulation phases such as exploration or grasping. Recognizing these phases allows for more adaptive and flexible control strategies suited to the specific demands of each phase \cite{fukui2012,veiga2020a}. Unique challenges arise in sequential (e.g., pick-and-lift) and interleaved phases (e.g., in-hand rolling), necessitating customized controllers. When similar phases recur, existing architectures can be redeployed, utilizing smooth blending strategies for phase transitions \cite{vien2015}.
        
        Given the unpredictable nature of tactile manipulation and its analytical complexity, data-driven strategies are typically employed \cite{lee2020,cui2020}. These strategies hinge on predictive techniques or learning from observed strategies to efficiently plan and execute manipulation tasks.

    \color{black}
    \section{Conclusion and Future Perspectives}
        \label{section:conclusion}
        Robotic manipulation tasks - continually evolving - rely heavily on tactile sensing to provide rich contact point data, thereby enhancing the system's understanding of physical interactions. Previous discussions have examined various control strategies, from active environment exploration and stable grasping to object manipulation. These strategies are distinguished by control type and information processing method, including information retrieval, analytical models, and AI.

        To enhance tactile manipulation further, it is vital to highlight key areas poised to improve performance and foster sophisticated interaction abilities. Three main areas emerge as promising avenues for this critical research domain.
        
        Firstly, the focus lies on sensor components. Crucial advancements include the development of hardware for new sensors that yield less noisy, more robust data about external stresses. Moreover, the need for a universal tactile sensor is apparent - one that offers comprehensive information without needing multiple separate sensors. Such a sensor would produce coherent, complete signals from measured interactions, providing advantages in physical manipulator integration and from a software perspective. Additionally, the rise of soft robotics presents opportunities to incorporate tactile sensing in deformable manipulators, enhancing their body awareness and versatility in manipulation capabilities.
        
        Secondly, attention must be given to modeling and learning techniques. While contributions to this field have seen significant growth, more development is needed, especially in learning. Advancements in learning algorithms can pave the way for more computationally efficient methods applicable directly on robots, bypassing extensive simulation needs. This growth is intertwined with addressing the stochastic nature of information from current tactile sensors and improving algorithms to manage such uncertainty effectively.
        
        Finally, advancements in control strategies play a crucial role. The task of enhancing control architectures in tactile manipulation is complex due to its close relationship with the previously mentioned factors. Nevertheless, it is imperative to develop control architectures considering the entire manipulation task, including often overlooked phenomena like manipulating deformable objects or collaborative tasks. These advanced control architectures should integrate tactile sensing information, use context clues from other sensory modalities, and support adaptive and robust manipulation.
        
        Achieving mastery over increasingly intelligent tactile manipulation promises to enhance performance and overcome tasks previously limited by technological or scientific barriers.

    \subsubsection*{Acknowledgments}
    This work was supported by the European Union’s Horizon 2020 Research and Innovation Program under Grant Agreement No. 863212 (PROBOSCIS project).

    \subsubsection*{Author contributions}
    E.D. organized the work and drafted the manuscript. E.D. and E.F. conceived and designed the survey process. M.L.P. contributed to draft the hardware section and to produce task illustrations. L.B. gave constructive advice for this work. E.F. directed, supervised and fully revised the research. All authors reviewed the manuscript.

    \subsubsection*{Competing interest}
    The authors declare no competing interests.
            
    \bibliography{references}

\end{document}